\documentclass[10pt,twocolumn,letterpaper]{article}

\usepackage[pagenumbers]{wacv} 

\usepackage{adjustbox}
\usepackage[dvipsnames]{xcolor}
\usepackage{graphicx}
\usepackage{multirow}
\usepackage{tabularx}

\usepackage{amsmath}
\usepackage{amssymb}
\usepackage{booktabs}
\newcolumntype{C}{>{\centering\arraybackslash}X}
\NewExpandableDocumentCommand\mcc{O{1}m}{\multicolumn{#1}{c}{#2}}

\newcommand{\nickname}{ST-CLIP}

\usepackage[pagebackref,breaklinks,colorlinks]{hyperref}

\usepackage[capitalize]{cleveref}
\crefname{section}{Sec.}{Secs.}
\Crefname{section}{Section}{Sections}
\Crefname{table}{Table}{Tables}
\crefname{table}{Tab.}{Tabs.}

\begin{document}

\title{Spatio-Temporal Context Prompting for Zero-Shot Action Detection}

\author{Wei-Jhe Huang$^1$ \quad
Min-Hung Chen$^2$ \quad
Shang-Hong Lai$^1$ \\
$^1$National Tsing Hua University, Taiwan \quad
$^2$NVIDIA \quad
}

\maketitle

\begin{abstract}
Spatio-temporal action detection encompasses the tasks of localizing and classifying individual actions within a video. Recent works aim to enhance this process by incorporating interaction modeling, which captures the relationship between people and their surrounding context. However, these approaches have primarily focused on fully-supervised learning, and the current limitation lies in the lack of generalization capability to recognize unseen action categories. In this paper, we aim to adapt the pretrained image-language models to detect unseen actions. To this end, we propose a method which can effectively leverage the rich knowledge of visual-language models to perform Person-Context Interaction. Meanwhile, our Context Prompting module will utilize contextual information to prompt labels, thereby enhancing the generation of more representative text features. Moreover, to address the challenge of recognizing distinct actions by multiple people at the same timestamp, we design the Interest Token Spotting mechanism which employs pretrained visual knowledge to find each person's interest context tokens, and then these tokens will be used for prompting to generate text features tailored to each individual. To evaluate the ability to detect unseen actions, we propose a comprehensive benchmark on J-HMDB, UCF101-24, and AVA datasets. The experiments show that our method achieves superior results compared to previous approaches and can be further extended to multi-action videos, bringing it closer to real-world applications. The code and data can be found in \href{https://webber2933.github.io/ST-CLIP-project-page/}{ST-CLIP}.
\end{abstract}
\section{Introduction}
\label{sec:intro}

\begin{figure}[t]
\begin{center}   \includegraphics[width=0.5\textwidth]{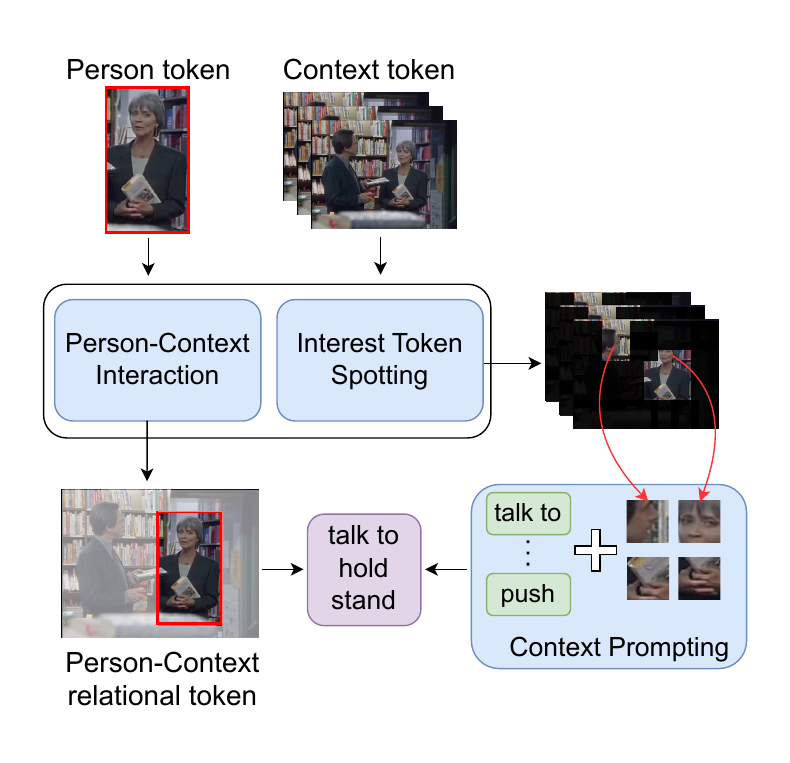}
\end{center}
   \caption{\textbf{Overview of our method.} We aim to transfer the knowledge of CLIP to detect unseen actions. We leverage the pretrained knowledge to model the interaction between people and their surrounding context. Besides, the Interest Token Spotting mechanism utilizes the knowledge to find the tokens most relevant to a person, then the Context Prompting uses these visual tokens to augment the text contents, which can make them easier to distinguish.}
\label{fig:teaser}
\end{figure}

The task of spatial-temporal action detection is to detect people and recognize their respective actions in both space and time, which holds broad applications in various fields, including self-driving cars, sports analysis, and surveillance. Recently, the rise of 3D CNN backbones \cite{Feichtenhofer_2019_ICCV,tran2015learning,tran2018closer} has strengthened the capabilities of representation learning in spatial-temporal context, which has greatly improved the performance of action detection. Furthermore, some recent studies have extended their focus by incorporating attention-based relation modeling \cite{tang2020asynchronous,pan2021actor,faure2023holistic}. These approaches aim to model the relationship between individuals and their surrounding environment, including other people, objects, and the contextual scene. By integrating more interaction information into the person feature, a more comprehensive representation of their actions is achieved, consequently enhancing the accuracy of action classification. However, these methods primarily center around fully supervised learning, limiting their capability to detect only the action classes included in the training phase. In real-world applications, numerous actions beyond the training classes are bound to occur. Therefore, our approach aims to push the boundaries further by detecting unseen actions in zero-shot scenarios, alleviating the considerable labor-intensive efforts associated with the annotation process.

In recent years, visual-language models \cite{radford2021learning,jia2021scaling,yuan2021florence} have gradually become the models of choice in zero-shot video understanding due to their strong generalization capability. Nevertheless, the predominant focus in this domain is currently on video classification \cite{wang2021actionclip,ni2022expanding,ju2022prompting,wasim2023vita,wu2023bidirectional} and temporal action detection \cite{ju2022prompting,nag2022zero}, where the entire video or a short clip is considered for action classification. \cite{huang2023interaction} is most similar to our goal, which is to detect individual unseen actions. However, the scenario they handle is too simple. They only process videos with single actions and target specific unseen labels, limiting their ability to effectively evaluate the robustness of the method. In contrast, our goal is to detect a variety of unseen actions and extend the method to videos containing multiple actions.

Towards the aforementioned goal, we propose a novel framework called \nickname, which adapts CLIP \cite{radford2021learning} to Zero-Shot Spatio-Temporal Action Detection in both visual and textual aspects, as shown in Figure \ref{fig:teaser}. In terms of vision, we propose to utilize the visual knowledge embedded in CLIP to perform Person-Context Interaction. This approach enables us to grasp the relationship between individuals and their surrounding context without the necessity for additional interaction modules, thereby preserving the generalization capabilities of CLIP and streamlining the interaction modeling process. In the textual domain, given that the class names in the dataset offer limited semantic information, the ambiguity between different labels may degrade the quality of the classification results. Our objective is to enhance the textual content through effective prompting. Inspired by \cite{ni2022expanding}, we design a multi-layer Context Prompting module, which incrementally utilizes visual clues from spatio-temporal context to augment text descriptions, thereby increasing the discrimination capability. Furthermore, given that real-world scenarios often involve multiple individuals concurrently performing different actions, we further introduce Interest Token Spotting, which aims to identify context tokens most relevant to each person's actions. Subsequently, these tokens are utilized in the prompting process to generate a description that aptly captures each individual's situation.

In order to assess the effectiveness of our method, we refer to \cite{huang2023interaction} and propose a more complete benchmark on J-HMDB, UCF101-24 and AVA datasets. For the first two datasets, we conduct cross-validation with varying train/test label combinations. This approach, as opposed to \cite{huang2023interaction}, which exclusively experiments with a specific label split, provides a more comprehensive assessment of the method's robustness. For experiments on AVA, where a single video may involve multiple actions, we randomly select certain videos that lack common classes for training, then the subsequent evaluation will focus on assessing the performance of detecting these unseen actions. The experimental results demonstrate that, in comparison to other zero-shot video classification methods, our approach exhibits superior performance on J-HMDB and achieves competitive results on UCF101-24. Furthermore, experiments on AVA demonstrate that our method can detect various unseen actions individually within the same video, affirming its potential extension to real-world applications. To summarize, our contributions are as follows:
\begin{itemize}
    \item
    We propose a novel method \nickname \space that fully leverages the visual-language model to capture the relationship between people and the spatial-temporal context, without training extra interaction modules.
    \item 
    We devise a multi-layer Context Prompting module that employs both low-level and high-level context information to prompt class names, enriching the semantic content. In addition, we introduce an Interest Token Spotting mechanism to identify tokens most relevant to individuals for prompting, thereby generating text features that are unique to each person.
    \item
    We propose a complete benchmark on J-HMDB, UCF101-24, and AVA datasets to evaluate performance on Zero-Shot Spatio-Temporal Action Detection. The experiments demonstrate the strong generalization capabilities of our method, and show the ability to individually detect unseen actions within the same video.
\end{itemize}

\section{Related Work}
\label{sec:Related}

\noindent\textbf{Spatio-Temporal Action Detection.} Typical action detection methods mostly use the two-stage pipeline, which means first localizing people in a video, and then performing action classification based on the features of these people. Most of these methods utilize additional person detectors like Faster R-CNN \cite{ren2015faster} to generate actor bounding boxes, which are used to perform RoIAlign \cite{he2017mask} on the video features generated by the 3D CNN backbone to obtain the person features. While \cite{Feichtenhofer_2019_ICCV} directly utilizes naive actor features to classify actions, \cite{faure2023holistic,tang2020asynchronous,pan2021actor} further exploit relation modeling to combine more information about human and environmental interactions. Besides, some methods \cite{sun2018actor,yang2019step,girdhar2019video,chen2021watch} optimize the two-stage networks by a joint loss in an end-to-end framework. There are also several methods designed to handle similar tasks in unsupervised scenarios. The goal of \cite{soomro2017unsupervised} is to localize the time and space of actions in the video without using bounding box annotations during training. \cite{agarwal2020unsupervised} propose a domain adaptation framework for action detection; however, their approach is limited to detecting fixed action classes, whereas our goal is to operate in an open-vocabulary setting.


\noindent\textbf{Video Understanding with Visual-Language Models.}
Recently, large-scale visual-language models such as CLIP \cite{radford2021learning}, ALIGN \cite{jia2021scaling}, and Florence \cite{yuan2021florence} have demonstrated their usability to different visual-language tasks including image captioning \cite{mokady2021clipcap}, video-text retrieval \cite{fang2021clip2video} and scene text detection \cite{yu2023turning}. Due to a shared feature space that effectively aligns the visual and text domains, an increasing number of methods \cite{wang2021actionclip,ju2022prompting,ni2022expanding,wasim2023vita,wu2023bidirectional} choose to perform zero-shot video classification based on these foundation models. The main focus of these works is to design temporal modeling to adapt the image encoder to the video domain and to develop ways to prompt the text. On this basis, \cite{huang2023interaction} processes zero-shot spatio-temporal action detection to further subdivide the scope of action classification to the individual level. They proposed extracting people and objects in the image and using different interaction blocks to model the relationship between them. Besides, they will use each person's interaction feature to prompt labels. To evaluate their performance, they selected specific unseen actions to detect on the two datasets, J-HMDB and UCF101-24. However, this benchmark is still not close enough to real-world scenarios. First, they did not extensively test a variety of unseen labels. Second, the videos in both datasets contain only single actions. Considering this, we propose a more complete benchmark to evaluate performance on multi-action videos, aiming for more practical applications.
\section{Proposed Method}
\subsection{Preliminary: Visual-Language Model}
As our method exploits the pretrained knowledge of the CLIP model, we briefly review the formulation of image and text encoders in this section. For the image encoder of ViT architecture \cite{dosovitskiy2020image}, given an image $I\in{\mathbb{R}^{H\times{W}\times{3}}}$ with height $H$ and width $W$, it will be split into $N = \frac{H}{P}\times\frac{W}{P}$ patches, where the patch size is $P\times{P}$, then each patch will obtain its token through patch embedding. In this process, the Conv2D with kernel size $P\times{P}$ and output channel size $D$ will be used to generate patch tokens $x\in{\mathbb{R}^{N\times{D}}}$. After that, an extra learnable token $x_{cls}\in\mathbb{R}^{D}$ is concatenated for classification. The input tokens for the transformer encoder layer is given by:
\begin{equation}
\begin{aligned}
& X = [x_{cls},x_1,x_2,\dots,x_N] + e
\end{aligned}
\end{equation}
where $e$ is the positional encoding. The classification token $x_{cls}$ output from the last encoder layer is often regarded as an image feature. Similarly, the text encoder is also a transformer architecture, while it first tokenizes the text into embeddings, and then uses the EOS token of the last encoder layer output as the text feature.

\begin{figure*}[t]
\begin{center}
   \includegraphics[width=1.0\textwidth]{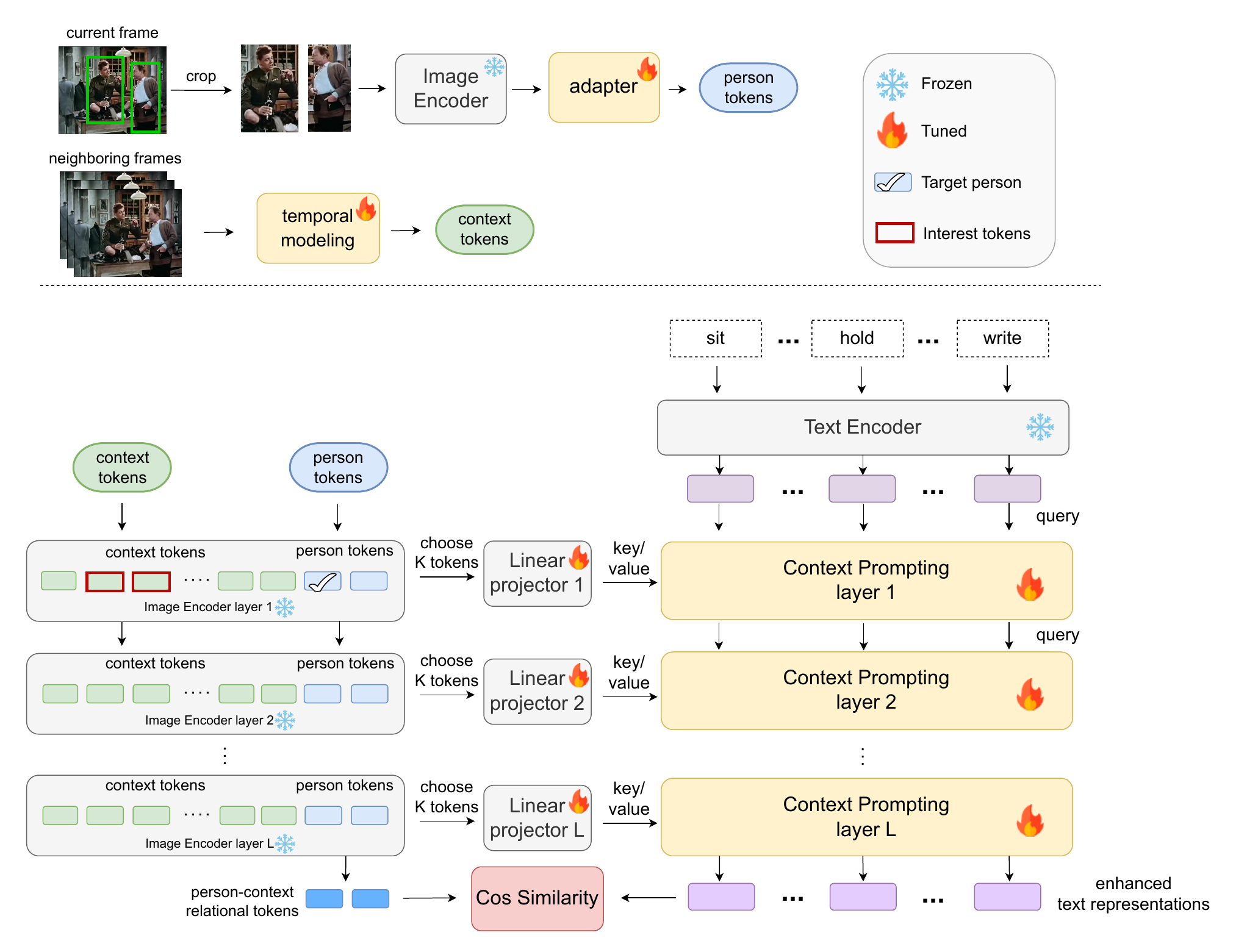}
\end{center}
   \caption{\textbf{\nickname \space framework.} We first extract the person tokens for the person bounding boxes detected from each frame. Then, we perform temporal modeling on the neighboring frames to obtain the context tokens. After that, we leverage the CLIP’s visual knowledge to perform person-context interaction on these tokens. In addition, we utilize the attention weight in each encoder layer to find the interest tokens for each person, then the Context Prompting layer will use these visual tokens to prompt the class names. Finally, the cosine similarities between person-context relational tokens and the label prompting features determine the classification scores for the actions. }
\label{fig:overview}
\end{figure*}

\subsection{Overall Architecture}
The overview of our \nickname \space is shown in Figure \ref{fig:overview}. As a two-stage framework, our model takes the person detected from video frames by a human detector as input and outputs the corresponding action classification results. Given an image with the person bounding boxes, we first extract these portions and obtain person-specific tokens through the image encoder. In this process, we utilize the adapter to make these person tokens more suitable for subsequent interaction modeling, which will be discussed later. Given the continuous nature of actions, in addition to considering a person’s information, we sample neighboring frames to construct a spatial-temporal context, thereby capturing information across different spaces and times. To obtain these context tokens, we conduct temporal modeling on the patch tokens of different frames, enabling the tokens to aggregate information over this period of time.

Subsequently, to fully leverage CLIP's visual knowledge, we jointly input person and context tokens into the image encoder. The Multi-Head Self Attention (MHSA) in each encoder layer is employed to guide us in achieving the following three objectives: \emph{(i)} performing further spatial modeling on the input context tokens to obtain spatial-temporal tokens. \emph{(ii)} modeling person-person and person-context interaction through the mutual influence between tokens. \emph{(iii)} identifying the interest tokens most relevant to each person's actions through attention weight. On the textual side, we initially utilize the CLIP text encoder to generate the original text features for class names. Then, followed by each image encoder layer, the Context Prompting layer will use context tokens to prompt each label. In this process, considering that the videos in J-HMDB and UCF101-24 only contain a single action, we can treat all context tokens as relevant to this action. Hence, we use all context tokens for prompting, resulting in everyone in the same frame sharing the same text features. However, in AVA, to discern different actions by multiple individuals, we utilize context tokens that each person deems important (\ie, interest tokens) to prompt their respective text features. Finally, we use the person tokens and label features output by the last image encoder layer and Context Prompting layer to calculate the cosine similarities, which are used as the action classification scores.

In the training process, in order to retain the generalization capability of CLIP for zero-shot tasks, we freeze the pretrained weights in the image and text encoders, and only train our additional learnable modules. Besides, we insert the LoRA trainable matrices into the Feed-Forward Network (FFN) of each image encoder layer, which can further adapt the CLIP model to detect actions without affecting its well-aligned visual-language features.

\subsection{Person-Context Interaction}
\begin{figure}[t]
\begin{center}   \includegraphics[width=0.5\textwidth]{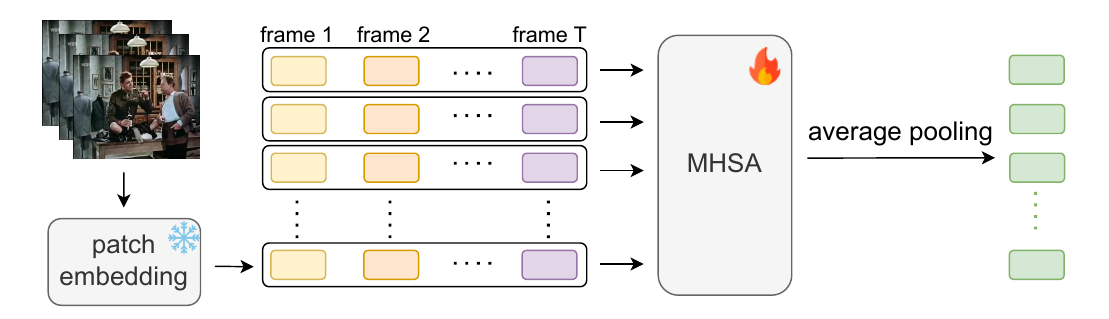}
\end{center}
   \caption{\textbf{Temporal modeling.} We apply self-attention along the temporal dimension to fuse the information.}
\label{fig:temporal_modeling}
\end{figure}

As mentioned earlier, to utilize spatial-temporal context and facilitate the recognition of continuous actions, we sample $T$ neighboring frames before and after the current frame. We first conduct temporal modeling on these frames to consolidate information at different times. Subsequently, we leverage the spatial modeling capability of the image encoder to further fuse these visual contents in both space and time, which results in the generation of spatial-temporal tokens. Our temporal modeling is shown in Figure \ref{fig:temporal_modeling}. First, we use CLIP's pretrained patch embedding to obtain the patch tokens of each frame $X_t = [x_{t,1},x_{t,2},\dots,x_{t,N}] \in{\mathbb{R}^{N\times{D}}}$, where $t \in \{1,\dots,T\}$ denotes the frame index, $N$ is the number of patch tokens, and $D$ is the token dimension. Then we gather the tokens of each frame into $[X_1,X_2,\dots,X_T] \in{\mathbb{R}^{T\times{N}\times{D}}}$. After that, we utilize MHSA to model the relationship between patch tokens at the same position in different frames $Z_i = [x_{1,i},x_{2,i},\dots,x_{T,i}] \in \mathbb{R}^{T\times{D}}$, where $i \in \{1,\dots,N\}$, as follows:
\begin{equation}
\begin{aligned}
& \Bar{Z_i} = Z_i + e^{temp} \\
& \hat{Z_i} = \Bar{Z_i} + MHSA(LN(\Bar{Z_i})) \\
& \Tilde{Z_i} = AvgPool(\hat{Z_i}),
\end{aligned}
\end{equation}
where $e^{temp}$ is the temporal encoding and $LN$ stands for layer normalization. After temporal modeling, we can obtain context tokens $\Tilde{Z_i} \in \mathbb{R}^D$ at each position $i \in \{1,\dots,N\}$ that have aggregated temporal information, which will be used to model the interaction with a person.

Regarding the person tokens, we initially utilize the image encoder to obtain features for each person. Considering that these person features have undergone multiple transformer encoder layers, they are relatively high-level compared to the aforementioned context tokens, which are only at the patch-embedding level. To enhance the utilization of self-attention in modeling relationships among all tokens, we employ an adapter to adjust person features to the same level as other context tokens. The adapter is a straightforward two-layer FFN commonly used in the transformer encoder layer, and we have found it to be effective in adapting these person features. Specifically, we generate person tokens in the following ways:
\begin{equation}
\begin{aligned}
& \Tilde{P_i} = P_i + FFN(LN(P_i))
\end{aligned}
\end{equation}
where $P_i$ is the person feature output from the image encoder and $\Tilde{P_i}$ is the person token we will use for subsequent interaction modeling.

\subsection{Interest Token Spotting}
In a multi-action dataset like AVA, the same timestamp may encompass various actions performed by multiple individuals, and all the context tokens will contain information about different actions. Therefore, we introduce Interest Token Spotting, a mechanism that employs a personal perspective to extract context tokens most relevant to each individual's actions. Subsequently, we utilize these tokens for prompting to generate personalized text features. In this process, we also leverage CLIP’s pretrained visual knowledge to find each person’s interest tokens. To be more specific, we exploit the attention weight calculated by the MHSA in each image encoder layer as an indicator of the token importance. The Multi-Head Self Attention will first calculate an attention map $M_i \in \mathbb{R}^{C\times{C}}$ based on the query and key of each head, where $i \in \{1,\dots,num\_heads\}$, and $C$ is the total number of person tokens and context tokens. Then, we average the attention maps of each head to obtain an importance score matrix $M \in \mathbb{R}^{C\times{C}}$. In this matrix, each row represents the importance of every token to a certain token. For instance, $M(i,j)$ represents how important token $j$ is to token $i$. Based on this, we use a person's token as the row index, and select the top $K$ highest among all $C$ importance scores. These selected $K$ indexes are the token positions that the person is interested in. After that, we pass each token through the MHSA and FFN of this encoder layer, and use the selected $K$ indexes to obtain the interest tokens.

\subsection{Context Prompting}

The Context Prompting layer primarily consists of Cross-Attention (CA), where text features serve as the query, and the context tokens act as key and value. This setup facilitates the gradual absorption of visual information into the text content. In AVA, we further narrow down the scope of the extracted information for each person to their personal interest tokens. Given an image with $B$ detected individuals, we need to classify their actions into $N_L$ possible labels. First, we assign the same set of original text features that are obtained by the CLIP text encoder to these $B$ people. Subsequently, we employ a linear projector to project each person's interest tokens to the same dimension as the text features. Following this, both the interest tokens and the text features are sent to the Context Prompting layer for prompting. The following equations describe how it works:
\begin{equation}
\begin{aligned}
    &\Bar{F_T} = {F_T} + CA(F_T, F_I),\\
    &\hat{F_T} = \Bar{F_T} + FFN(\Bar{F_T}),\\
    & \Tilde{F_T} = F_T + \rho\hat{F_T},
\end{aligned}
\end{equation}
where $F_T \in \mathbb{R}^{B\times{N_L}\times{D}}$ consists of the text features, $F_I \in \mathbb{R}^{B\times{K}\times{D}}$ are the interest tokens, $\rho \in \mathbb{R}^D$ is a learnable weight vector, and $\Tilde{F_T}$ will be input to the next prompting layer. For J-HMDB and UCF101-24, we straightforwardly use all context tokens for prompting, making all $B$ individuals in an image utilize the same text features $F_T \in \mathbb{R}^{{N_L}\times{D}}$ for similarity calculation.

\section{Datasets and Benchmarks}



We establish benchmarks for zero-shot spatial-temporal action detection on three popular datasets: J-HMDB, UCF101-24, and AVA. For the first two datasets, we further extend the settings of \cite{huang2023interaction} to include more diverse unseen actions. Besides, we also build benchmark on AVA which is more representative of real-world scenarios. We use frame mAP with 0.5 IoU threshold in all the benchmarks for evaluation. More details about the label splits are described in the supplementary materials.

\subsection{ZS-JHMDB}
J-HMDB dataset \cite{jhuang2013towards} is a subset of the HMDB51 dataset. It has 21 classes and 928 videos. The videos are trimmed and there are totally 31,838 annotated frames in these videos. To assess the generalization capability of an action detection method, the zero-shot evaluation necessitates that the model has not seen samples related to test classes during the training process, which means the training and testing labels are disjoint. In this scenario, we refer to the evaluation settings proposed by \cite{huang2023interaction}, which exploits random sampling to take 75\% action classes for training, and the remaining 25\% for testing. However, they only employ a specific label split for evaluation, which is inadequate for fully measuring the effectiveness of the method. Instead, we perform cross-validation on multiple label splits as follows: we split J-HMDB into 4 label splits, each split has 15 training classes and 6 testing classes, and the testing classes in each split are disjoint (part of split 4 will overlap with split 1). The split 1 is the same as the split used in \cite{huang2023interaction}.

\subsection{ZS-UCF}
UCF101-24 dataset is a subset of UCF101 \cite{soomro2012ucf101}. It consists of 3207 videos from 24 action classes, and each video contains a single action. In this benchmark, we employ the same setting as in ZS-JHMDB, which also divides all classes into 75\% for training and 25\% for evaluation. The label split 1 is also the same as used in \cite{huang2023interaction}. 

\subsection{ZS-AVA}
AVA\cite{gu2018ava} is a large-scale action detection dataset that contains multiple actions within a single video. It consists of 235 videos for training and 64 videos for validation. Each video lasts for 15 minutes and is annotated by sampling one keyframe per second. For AVA, since the same video contains multiple actions, and some action categories exist in multiple videos, it becomes challenging to find a sufficient amount of training data and testing data with disjoint labels. Instead, we randomly select some training videos, ensuring that they all lack samples of the same classes. These missing classes are then treated as unseen classes for evaluation. During the evaluation phase, we test all classes in the validation videos, but the focus is solely on evaluating the performance of unseen classes. Under this setting, we propose three splits. When the amounts of training and testing data for these splits are nearly the same, we select different combinations of three action types — pose action, object interaction, and person interaction as unseen classes, allowing for a more comprehensive evaluation.
\section{Experiments}
\subsection{Experimental Setup}
\noindent\textbf{Person Detector:} In the following experiments, we employ Faster R-CNN \cite{li2021benchmarking} with a ResNet-50-FPN \cite{lin2017feature} backbone pretrained on MSCOCO \cite{lin2014microsoft} for person detection. For J-HMDB and AVA, we directly inference on the test data with pretrained person detector. For UCF101-24, since the images have lower resolution, we further use the ground truth of training classes to finetune the person detector for 10 epochs in each label split. Besides, to remove the false positives, we keep the detected box with the highest confident score $S$ in each frame, then we select the boxes with scores higher than $S-T$ from the rest, where $T$ is 0.001 for J-HMDB and 0.7 for both UCF101-24 and AVA.

\noindent\textbf{Hyperparameters:} For all the experiments on J-HMDB and UCF101-24, we employ ViT-B/16 as our CLIP backbone and use the same hyper-parameters as follows: Training for 3K iterations with a batch size of 8. We use SGD as our optimizer and the base learning rate is set to 2.5e-4. As for AVA, we use the ViT-L/14 backbone and train the model for 20K iterations with a base learning rate of 4e-4.

\subsection{Compared Methods}
In addition to comparing our approach with iCLIP \cite{huang2023interaction}, which addresses the same task as ours, we also evaluate it against the following methods.

\noindent\textbf{Baseline:} 
we follow \cite{huang2023interaction} to implement a naive baseline for comparison. For a frame with detected individuals, the baseline utilizes the pretrained image encoder of CLIP to extract the image feature of this frame. Subsequently, it calculates the cosine similarities with the text features of each class name, which are then considered as the action classification scores for these individuals. Since the baseline regards people in the same frame as having the same actions, we also implement the person crop method for multi-action videos. This involves cropping out parts of each person to obtain their respective image features for classification.

\noindent\textbf{ViCLIP \cite{wang2023internvid}:} We also experiment with a video-language model for individual action classification. First, we extract the video feature map using a video encoder and apply average pooling along the temporal dimension. Then, for each person, we use their bounding box to perform ROIAlign and max pooling on the feature map, obtaining a person-specific feature for classification.

\noindent\textbf{Video classification methods \cite{wang2021actionclip, ju2022prompting, ni2022expanding, wasim2023vita}:} since each video in both ZS-JHMDB and ZS-UCF contains only a single action, a straightforward approach is to conduct action detection through video classification. These methods can initially classify the entire video into an action class and then consider all detected individuals in the video as performing this action. As for ZS-AVA, the main distinction between our method and these approaches is our capability to detect different actions within the same video, which is difficult to achieve. Firstly, in the training process of these methods, providing a fixed video label is challenging due to the presence of numerous different actions. Additionally, during the inference stage, these methods tend to detect that everyone in the video has the same action. To study the feasibility of these methods, we further narrow the scope of classification from the entire video to tracklets to avoid misclassifying different actions into the same category. 

\subsection{Zero-Shot Spatial-Temporal Action Detection}

\begin{table}[ht]
\begin{adjustbox}{center}
\setlength{\tabcolsep}{1mm}
\scalebox{0.82}{
\begin{tabular}{lccccccc}
    \toprule
    & \multicolumn{7}{c}{Frame mAP@0.5} \\
    \cmidrule(l){2-8} 
    \multirow{-2}{*}{Method} & split 1 & split 2 & split 3 & avg & H & avg* & H*\\
    \midrule
    baseline & 8.67 & 8.57 & 2.77 & 6.67 & 7.38 & 10.09 & 10.58\\
    baseline (person crop) & 8.22 & 5.05 & 3.04 & 5.44 & 6.42 & 7.55 & 8.63\\
    iCLIP \cite{huang2023interaction} & 4.04 & 9.08 & 1.91 & 5.01 & 7.59 & 6.91 & 10.37\\
    Vita-CLIP \cite{wasim2023vita} & 4.25 & 4.29 & 0.64 & 3.06 & 4.49 & 10.45 & 13.94\\
    \textbf{\nickname \space (Ours)} & \textbf{12.85} & \textbf{10.17} & \textbf{4.01} & \textbf{9.01} & \textbf{11.41} & \textbf{11.76} & \textbf{15.05}\\
    \bottomrule
\end{tabular}}
\end{adjustbox}
\caption{\textbf{Evaluation on ZS-AVA.} * denotes using the ground-truth bounding boxes of the test data. All methods employ the ViT-L/14 backbone. In the inference stage of Vita-CLIP \cite{wasim2023vita}, we use two different tracklets: (1) tracklets obtained by associating detected boxes with ByteTrack \cite{zhang2022bytetrack}, and (2) tracklets with ground-truth boxes provided by AVA official.}
\label{tab:ava_evaluation}
\end{table}

In this section, we present the experimental results for unseen classes in each label split. We also calculate the harmonic mean (H) of the average performance of both base and unseen classes, to provide a more comprehensive assessment. In addition to using detected person boxes to measure the actual results on zero-shot spatial-temporal action detection, we also provide results using ground-truth bounding boxes of the test data. This allows us to analyze the performance without the influence of localization errors. 

Table \ref{tab:ava_evaluation} shows our results on detecting unseen actions in AVA. First of all, the baseline method aligns image features during both the pretraining phase of CLIP and the inference phase of action detection, leading to good performance on unseen classes. It is worth noting that although iCLIP \cite{huang2023interaction} and the baseline method achieve similar harmonic mean, iCLIP's performance on unseen classes is relatively poor. This suggests that their interaction modeling affects generalization ability. In contrast, our Person-Context Interaction approach leverages CLIP's pretrained knowledge, resulting in superior performance on both base and unseen classes.

We present the experimental results on ZS-JHMDB and ZS-UCF in \cref{tab:jhmdb7525,tab:ucf7525}. Since our method focuses on detecting different actions for each person, individuals in the same video may be classified into different actions. This general setting can result in other video classification methods having an advantage over us in these two datasets. For a fair comparison with the other methods, we further adopt the assumption that a video contains only one action. In this context, we perform soft voting on each person's classification score, extending our method to suit this scenario.


We first present the results on ZS-JHMDB in Table \ref{tab:jhmdb7525}. Firstly, the video-language model \cite{wang2023internvid} with ROI align has difficulty accurately detecting unseen actions. We speculate that this is primarily due to the alignment of global video features (CLS token) during the pretraining process of ViCLIP. As a result, the person-specific features extracted from the video feature map could not align well with the text features.
Without the assumption, our \nickname \space achieves the highest average performance on unseen classes, along with the best harmonic mean. With the assumption applied, our method still outperforms other video classification approaches, achieving an average performance 3.82 mAP higher than \cite{wasim2023vita}. Furthermore, by using ground-truth bounding boxes to eliminate localization errors, our performance improves further to 90.12 mAP.

\begin{table}[ht]
\begin{adjustbox}{center}
\setlength{\tabcolsep}{1mm}
\scalebox{0.82}{
\begin{tabular}{lcccccccc}
    \toprule
    & \multicolumn{8}{c}{Frame mAP@0.5} \\
    \cmidrule(l){2-9} 
    \multirow{-2}{*}{Method} & split 1 & split 2 & split 3 & split 4 & avg & H & avg* & H*\\
    \midrule
    \multicolumn{9}{l}{\textit{Without the assumption of single-action video}}
    \\
    baseline & 64.63 & 71.04 & 82.13 & 75.88 & 73.42 & 65.45 & 81.21 & 71.82\\
    ViCLIP \cite{wang2023internvid} & 52.29 & 72.29 & 53.59 & 70.58 & 62.19 & 68.03 & 68.15 & 74.05 \\
    iCLIP \cite{huang2023interaction} & 66.53 & 69.99 & 82.88 & 71.84 & 72.81 & 74.47 & 79.01 & 81.09\\
    \textbf{\nickname \space (Ours)} & \textbf{74.55} & \textbf{74.97} & \textbf{83.59} & \textbf{83.22} & \textbf{79.08} & \textbf{77.65} & \textbf{85.53} & \textbf{84.32}\\
    \midrule
    \multicolumn{9}{l}{\textit{With the assumption of single-action video}} \\
    ActionCLIP \cite{wang2021actionclip} & 69.18 & 75.28 & 77.11 & 76.55 & 74.53 & 75.82 & 82.07 & 83.25\\
    A5 \cite{ju2022prompting} & 50.92 & 66.06 & 69.07 & 60.21 & 61.57 & 68.94 & 67.39 & 75.34\\
    X-CLIP \cite{ni2022expanding} & 72.91 & 72.62 & 80.02 & 77.78 & 75.83 & 77.13 & 83.11 & 84.34\\
    Vita-CLIP \cite{wasim2023vita} & 68.60 & \textbf{82.35} & 84.95 & 80.19 & 79.02 & 80.46 & 86.02 & 87.49\\
    \textbf{\nickname \space (Ours)} & \textbf{79.62} & 78.70 & \textbf{85.84} & \textbf{87.19} & \textbf{82.84} & \textbf{80.96} & \textbf{90.12} & \textbf{88.48}\\
    \bottomrule
\end{tabular}}
\end{adjustbox}
\caption{\textbf{Evaluation on ZS-JHMDB.} * denotes using the ground-truth boxes of the test data. All methods use the ViT-B/16 backbone.}
\label{tab:jhmdb7525}
\vspace*{-\baselineskip}
\end{table}

\begin{table}[ht]
\begin{adjustbox}{center}
\setlength{\tabcolsep}{1mm}
\scalebox{0.82}{
\begin{tabular}{lcccccccc}
    \toprule
    & \multicolumn{8}{c}{Frame mAP@0.5} \\
    \cmidrule(l){2-9} 
    \multirow{-2}{*}{Method} & split 1 & split 2 & split 3 & split 4 & avg & H & avg* & H*\\
    \midrule
    \multicolumn{9}{l}{\textit{Without the assumption of single-action video}} \\
    baseline & 48.37 & 52.84 & 39.76 & 51.92 & 48.22 & 49.00 & 90.49 & 82.45 \\
    ViCLIP \cite{wang2023internvid} & 41.37 & 43.93 & 24.14 & 34.04 & 35.87 & 43.72 & 63.07 & 72.50\\
    iCLIP \cite{huang2023interaction} & \textbf{50.34} & 52.75 & 39.73 & 47.95 & 47.69 & \textbf{54.46} & 85.47 & 90.35 \\
    \textbf{\nickname \space (Ours)} & 49.09 & \textbf{54.95} & \textbf{42.05} & \textbf{53.92} & \textbf{50.00} & 54.41 & \textbf{91.80} & \textbf{91.34} \\
    \midrule
    \multicolumn{9}{l}{\textit{With the assumption of single-action video}} \\
    ActionCLIP \cite{wang2021actionclip} & \textbf{52.64} & 55.01 & 38.04 & 51.84 & 49.38 & 55.36 & 89.72 & 92.91\\
    A5 \cite{ju2022prompting} & 46.78 & 47.36 & 41.23 & 53.18 & 47.14 & 54.24 & 85.60 & 90.86\\
    X-CLIP \cite{ni2022expanding} & 50.10 & 56.94 & 44.52 & 55.35 & \underline{51.73} & 56.96 & 94.12 & 95.30\\
    Vita-CLIP \cite{wasim2023vita} & 52.52 & \textbf{57.22} & \textbf{45.12} & \textbf{57.32} & \textbf{53.05} & \textbf{57.65} & \textbf{96.57} & \textbf{95.94}\\
    \textbf{\nickname \space (Ours)} & 51.36 & 56.30 & 43.12 & 54.52 & 51.33 & 55.94 & \underline{95.02} & 94.75 \\
    \bottomrule
\end{tabular}}
\end{adjustbox}
\caption{\textbf{Evaluation on ZS-UCF.} * denotes using the ground-truth boxes of the test data. All methods use the ViT-B/16 backbone.}
\label{tab:ucf7525}
\end{table}


Table \ref{tab:ucf7525} presents the results on ZS-UCF. 
It is worth mentioning that the localization errors have a noticeable impact on our performance in this dataset. Since there are instances where irrelevant people who are not performing actions are detected, these individuals should be considered as part of the background. However, these false positive cases will also contribute to the soft voting process, leading to our classification results being slightly inferior to other methods that solely rely on sampled frames to determine video labels. In this case, our method still outperforms \cite{wang2021actionclip,ju2022prompting}, and exhibits similar performance to \cite{ni2022expanding}. With ground-truth bounding boxes, our method achieves the second-best average performance. However, it is important to note that the UCF101-24 dataset has relatively low image quality, with most characters occupying only a small portion of the frame. This impacts the accuracy of our individual classification, putting us at a slight disadvantage compared to \cite{wasim2023vita}, which relies on the entire sampled frame.

\subsection{Ablation Study}
We present an ablation study in \cref{tab:Proposed components,tab:Comparison with iCLIP,tab:Context prompting,tab:Person tokens,tab:Context tokens} to investigate different design choices in our method. The experiments are performed on the label split 1 of ZS-JHMDB and ZS-AVA.

\noindent\textbf{Proposed components:} We first investigate the importance of each component in Table \ref{tab:Proposed components}. On J-HMDB, our proposed Person-Context Interaction effectively models the relationship between individuals and their surroundings, resulting in a 2.35 mAP improvement compared to the baseline. Furthermore, the Context Prompting module leverages context information to enhance text content, leading to an additional improvement of 7.57 mAP. On AVA, the above two components also demonstrate their effectiveness. Additionally, in multi-action videos, our Interest Token Spotting can identify context tokens most relevant to individual actions for prompting, further enhancing performance.
\begin{table}[ht]
\begin{adjustbox}{center}
\begin{tabular}[t]{lcc}
    \toprule
    Components & J-HMDB & AVA \\
    \midrule
    Baseline & 64.63 & 8.67 \\
    + Person-Context Interaction & 66.98 & 10.41 \\
    + Context Prompting & \textbf{74.55} & 12.07 \\
    + Interest Token Spotting & - & \textbf{12.85} \\
    \bottomrule
  \end{tabular}
\end{adjustbox}
\caption{\textbf{Proposed components}}\label{tab:Proposed components}
\end{table}

\noindent\textbf{Comparison with iCLIP \cite{huang2023interaction}:} We demonstrate the advantages of our method compared to \cite{huang2023interaction} in Table \ref{tab:Comparison with iCLIP}. Without prompting, our method performs slightly better than \cite{huang2023interaction}, and we do not need to use additional object detectors and interaction blocks in the interaction modeling process. In addition, our prompting strategy uses multiple levels of context tokens to augment text content, resulting in an improvement of 7.57 mAP. However, the prompting method of \cite{huang2023interaction} relies heavily on the results of interaction modeling, which limits their performance improvement.\newline

\noindent\textbf{Context prompting:} The results in Table \ref{tab:Context prompting} show the effectiveness of our prompting strategy. The results demonstrate that our prompting method, which utilizes tokens from low-level to high-level, yields better outcomes compared to using only high-level tokens from the last layer.\newline

\begin{table}[ht]
    \small
\begin{tabularx}{\linewidth}{@{} *{2}{C} @{}}
    \begin{tabular}[t]{@{} lr @{}}
    \toprule
    Method & + prompt \\
    \midrule
    iCLIP \cite{huang2023interaction} & 66.02 $\xrightarrow{{\color{teal}\textbf{+0.51}}}$ 66.53 \\
    Ours & \textbf{66.98} $\xrightarrow{{\color{teal}\textbf{+7.57}}}$ \textbf{74.55} \\
    \bottomrule
    \end{tabular}
    \caption{\textbf{Comparison with iCLIP}}\label{tab:Comparison with iCLIP} & 
    \begin{tabular}[t]{@{} lc @{}}
    \toprule
    Prompting & mAP \\
    \midrule
    w/o prompting & 66.98 \\
    only in last layer & 66.02 \\
    in every layer & \textbf{74.55} \\
    \bottomrule
    \end{tabular}
    \caption{\textbf{Context prompting}}\label{tab:Context prompting}
\end{tabularx}
\vspace*{-\baselineskip}
\vspace*{-\baselineskip}
    \end{table}

\noindent\textbf{Person tokens:} Table \ref{tab:Person tokens} explores different ways of generating person tokens. Initially, the simple approach of pooling over the embeddings of all patches in the person crop fails to deliver satisfactory performance. Furthermore, equipping adapters can perform better than using person features at the image encoder level. This demonstrates that our adapter can effectively adapt person tokens, making them more suitable for the subsequent person-context interaction.\newline

\noindent\textbf{Context tokens:} Table \ref{tab:Context tokens} shows the importance of temporal modeling. Using only the current frame to obtain context tokens results in the worst performance, indicating that aggregating temporal information is beneficial when identifying continuous actions. Additionally, equipping only a 1-layer MHSA can significantly improve performance compared to simple average pooling.

\begin{table}[ht]
    \small
\begin{tabularx}{\linewidth}{@{} *{2}{C} @{}}
    \begin{tabular}[t]{@{} lc @{}}
    \toprule
    Person tokens & mAP \\
    \midrule
    \textit{Without the adapter} \\
    Patch-embedding & 70.92 \\
    Image encoder & 70.29 \\
    \midrule
    \textit{With the adapter} \\
    1-layer FC & 67.37 \\
    2-layer FFN & \textbf{74.55} \\
    \bottomrule
    \end{tabular}
    \caption{\textbf{Person tokens}}\label{tab:Person tokens} & 
    \begin{tabular}[t]{@{} lc @{}}
    \toprule
    Temporal modeling & mAP \\
    \midrule
    w/o temporal & 68.03 \\
    average pooling & 68.63 \\
    1-layer MHSA & \textbf{74.55} \\
    2-layer MHSA & 72.06 \\
    \bottomrule
    \end{tabular}
    \caption{\textbf{Context tokens}}\label{tab:Context tokens}
\end{tabularx}
    \end{table}

\vspace*{-\baselineskip}
\vspace*{-\baselineskip}
\section{Conclusion}
In this paper, we explore zero-shot spatio-temporal action detection. We propose a complete benchmark on J-HMDB, UCF101-24, and AVA. Besides, we propose a method to adapt the visual-language model for this task. The Person-Context Interaction employs pretrained knowledge to model the relationship between people and their surroundings, and the Context Prompting module utilizes visual information to augment the text content. To address multi-action videos, we further introduce the Interest Token Spotting mechanism to identify the visual tokens most relevant to each individual action. The experiments demonstrate that our method achieves competitive performance compared to other video classification methods and can also handle multi-action videos.

\noindent{\textbf{Acknowledgement:}} This work was supported by NVIDIA Taiwan Research \& Development Center (TRDC). 
\setcounter{section}{0}
\renewcommand*{\theHsection}{chX.\the\value{section}}
\begin{center}
\textbf{\large Supplementary Material}
\end{center}
In the supplementary material, we present additional experimental results to substantiate the efficacy of our method and provide more details on the experimental settings. Initially, we elaborate our proposed benchmark for Zero-Shot Spatio-Temporal Action Detection in Sec. \ref{sec:Label Splits Details}. Subsequently, we showcase a visualization depicting interest tokens on ZS-AVA in Sec. \ref{sec:Interest Tokens}. Then, in Sec. \ref{sec:Text Features Distribution}, we illustrate the distribution of text features on both ZS-JHMDB and ZS-UCF to assess the impact of prompting. We then provide the complexity analysis of our method and others in Sec. \ref{sec:Complexity Analysis}. Besides, we provide more experimental analysis in Sec. \ref{sec:Additional Experimental Analysis} and describe how we handle multi-label prediction in Sec. \ref{sec:Multi-Label Prediction on AVA}. We also present the results using ground-truth bounding boxes on each benchmark in Sec. \ref{sec:Results with Groundtruth Bounding Boxes}. Finally, we discuss some limitations of our approach in Sec. \ref{sec:Limitations}, and give the implementation details of other methods in Sec. \ref{sec:Implementation Details}.

\section{Label Splits Details}
\label{sec:Label Splits Details}

In this section, we provide details of each label split used in our benchmarks. For ZS-JHMDB and ZS-UCF, we perform cross-validation on 4 label splits to assess the efficacy of our method. Each label split of ZS-JHMDB has 15 training classes and 6 testing classes, and each split of ZS-UCF has 18 training classes and 6 testing classes. More specifically, within each label split, there are 6 classes for testing, and all the remaining classes in the dataset are designated as training classes. In each label split, we follow the official split 1 of the two datasets, J-HMDB and UCF101-24, obtaining the training videos of the training classes and the testing videos of the test classes.

For ZS-AVA, to ensure an adequate volume of training data, we refrain from utilizing classes that frequently appear in most training videos as unseen classes. Our objective is to diversify each split by incorporating various types of unseen classes, including pose actions, object interactions, and person interactions. The split 1 contains 5 pose actions and 13 object interactions, split 2 contains 2 pose actions, 6 object interactions and 4 person interactions, and the split 3 contains 5 object interactions and 1 person interactions. The testing classes in each label split are shown in Table \ref{tab:label_split_jhmdb_ucf_ava}. 

\begin{table*}
\begin{adjustbox}{center}
\begin{tabularx}{\textwidth}{XXXXXXX}
    \hline
    \hline
    ZS-JHMDB \\
    \hline
    \hline
    split 1 & clap & sit & wave & throw & pullup & catch \\
    \midrule
    split 2 & kick\newline ball & run & climb \newline stairs & stand & shoot \newline gun & pick \\
    \midrule
    split 3 & brush \newline hair & push & pour & shoot \newline bow & jump & shoot \newline ball \\
    \midrule
    split 4 & golf & swing\newline baseball & walk & clap & sit & wave \\
    \hline
    \hline
    ZS-UCF \\
    \hline
    \hline
    split 1 & Ice\newline Dancing & Floor\newline Gymnastics & Salsa\newline Spin & Skate\newline Boarding & Soccer\newline Juggling & Volleyball\newline Spiking \\
    \midrule
    split 2 & Basketball & Skiing & Biking & Golf\newline Swing & Cliff\newline Diving & Diving \\
    \midrule
    split 3 & Fencing & Horse\newline Riding & Surfing & Long\newline Jump & Pole\newline Vault & Basketball\newline Dunk \\
    \midrule
    split 4 & Rope\newline Climbing & Skijet & Tennis\newline Swing & Trampoline\newline Jumping & Cricket\newline Bowling & Walking\newline With\newline Dog \\
    \hline
    \hline
    ZS-AVA \\
    \hline
    \hline
    split 1 & brush\newline teeth \newline \newline extract \newline \newline row\newline boat& chop\newline \newline \newline fishing \newline \newline sail\newline boat & cook\newline \newline \newline jump/\newline leap \newline \newline shovel & crawl\newline \newline \newline kick\newline (an object) \newline \newline stir & dance\newline \newline \newline martial art \newline \newline swim & play\newline board game \newline \newline dig \newline \newline take\newline a photo \\
    \midrule
    split 2 & hand\newline clap \newline \newline kick\newline (a person) & lift\newline (a person)\newline \newline martial\newline art & play\newline board game\newline \newline play\newline with kids & swim\newline \newline \newline row\newline boat & shovel\newline \newline \newline play\newline with pets & take\newline a photo \newline \newline work on\newline a computer \\
    
    \midrule
    split 3 & hug\newline (a person) & press & shovel & stir & text on/look at a\newline cellphone & turn\newline (e.g., a\newline screwdriver) \\
    \bottomrule
\end{tabularx}
\end{adjustbox}
\caption{\textbf{Testing classes in each label split on ZS-JHMDB, ZS-UCF and ZS-AVA.}}
\label{tab:label_split_jhmdb_ucf_ava}
\end{table*}

\section{Interest Tokens}
\label{sec:Interest Tokens}

\begin{figure*}[ht]
\begin{subfigure}{1.0\textwidth}
   \includegraphics[width=1.0\textwidth]{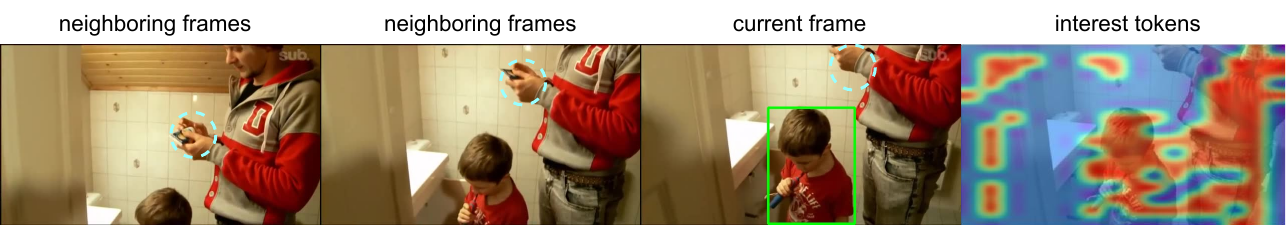}
   \caption{\textbf{Unseen action: brush teeth.} Compared with utilizing all context tokens for prompting, the use of only interest tokens increases the confidence score from 0.34 to 0.45.}
\label{fig:interest_tokens_brush_teeth}
\end{subfigure} \\

\begin{subfigure}{1.0\textwidth}
   \includegraphics[width=1.0\textwidth]{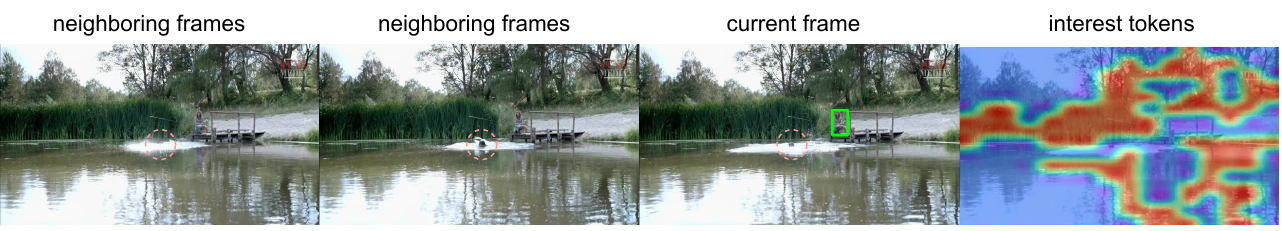}
   \caption{\textbf{Unseen action: fishing.} Compared with utilizing all context tokens for prompting, the use of only interest tokens increases the confidence score from 0.30 to 0.34.}
\label{fig:interest_tokens_fishing}
\end{subfigure}
\caption{\textbf{The impact of interest tokens.}}
\label{Fig:interest_tokens_with_neighbor}
\end{figure*}

We first explore the effects of utilizing only interest tokens to prompt labels, as opposed to incorporating all context tokens. In Figure \ref{Fig:interest_tokens_with_neighbor}, we showcase the current frame along with a bounding box, indicating our objective of detecting the person's action. Additionally, we include neighboring frames to facilitate the observation of changes in the video over this period. Since the context tokens have not undergone spatial modeling in the first image encoder layer, we choose to visualize the results obtained from this layer to improve the interpretability of the identified interest tokens. In Figure \ref{fig:interest_tokens_brush_teeth}, the target person is engaged in the action of brushing teeth, while another person undertaking distinct actions is highlighted within the circle. In this scenario, our Interest Token Spotting mechanism can identify the context tokens most relevant to the action of brushing teeth. This ensures that the prompting process selectively incorporates these crucial visual clues, enhancing the focus on pertinent information. Likewise, in Figure \ref{fig:interest_tokens_fishing}, the individual highlighted as the target person is engaged in fishing, while the person within the circled area is swimming. In this instance, the identified interest tokens exclude information associated with swimming, consequently enhancing the quality of the prompting process. Furthermore, in comparison to utilizing all context tokens for prompting, the use of only interest tokens can elevate the confidence score for the "brush teeth" action from 0.34 to 0.45, and the score of "fishing" can also be improved from 0.30 to 0.34. The results demonstrate that even when dealing with an unseen action not encountered during the training process, our method excels in identifying information most relevant to this action within a multi-person environment.

\begin{figure*}
\begin{subfigure}{0.5\textwidth}
   \includegraphics[width=1.0\textwidth]{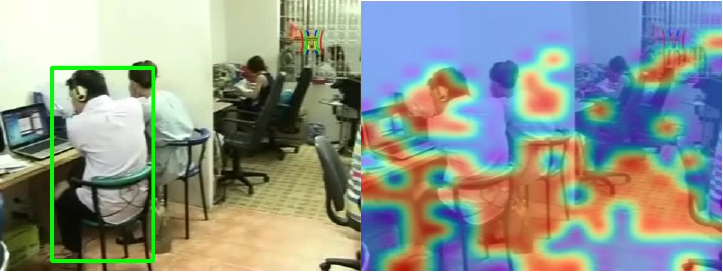}
   \caption{sit, \textbf{work on a computer} \ }
\label{fig:interest_tokens_work_on_a_computer}
\end{subfigure}
\begin{subfigure}{0.5\textwidth}
   \includegraphics[width=1.0\textwidth]{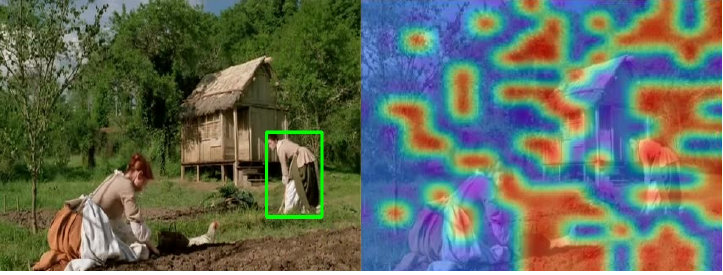}
   \caption{bend/bow (at the waist), carry/hold (an object), \textbf{shovel}}
\label{fig:interest_tokens_shovel}
\end{subfigure}
\begin{subfigure}{0.5\textwidth}
   \includegraphics[width=1.0\textwidth]{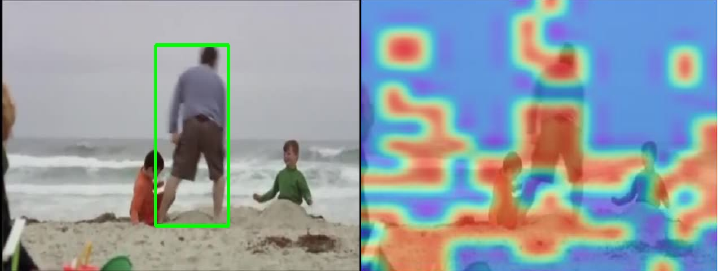}
   \caption{bend/bow (at the waist), watch (a person), \textbf{lift (a person)}, \textbf{play with kids}}
\label{fig:interest_tokens_play_with_kids}
\end{subfigure}
\begin{subfigure}{0.5\textwidth}
   \includegraphics[width=1.0\textwidth]{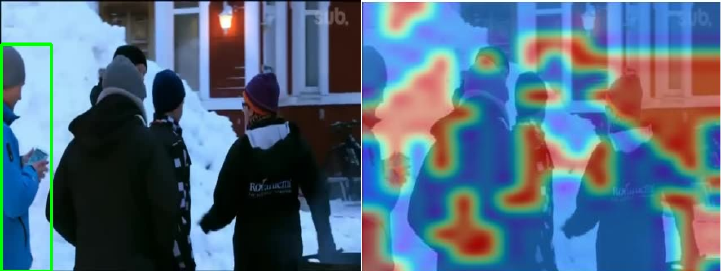}
   \caption{walk, listen to (a person), \textbf{text on/look at a cellphone}}
\label{fig:interest_tokens_look_at_a_cellphone}
\end{subfigure}
\caption{\textbf{More visualization of interest tokens.} Bold text indicate unseen actions.}
\label{Fig:interest_tokens}
\end{figure*}

We provide more visualization results in Figure \ref{Fig:interest_tokens}. In Figure \ref{fig:interest_tokens_work_on_a_computer} and \ref{fig:interest_tokens_shovel}, when the action performed by the target person has limited relevance to others, the identified interest tokens will exhibit reduced emphasis on areas involving other people. Conversely, in Figure \ref{fig:interest_tokens_play_with_kids} and \ref{fig:interest_tokens_look_at_a_cellphone}, where the person's actions interact with others, our method adeptly recognizes context tokens within other people's areas as interest tokens, effectively extracting relevant information from those regions. Moreover, our Interest Token Spotting possess the capability to identify crucial objects, enhancing our ability to recognize actions, such as the chair and the computer in Figure \ref{fig:interest_tokens_work_on_a_computer}, and the cellphone in Figure \ref{fig:interest_tokens_look_at_a_cellphone}.

\section{Text Features Distribution}
\label{sec:Text Features Distribution}

\begin{figure*}
\begin{subfigure}{1.0\textwidth}
   \includegraphics[width=1.0\textwidth]{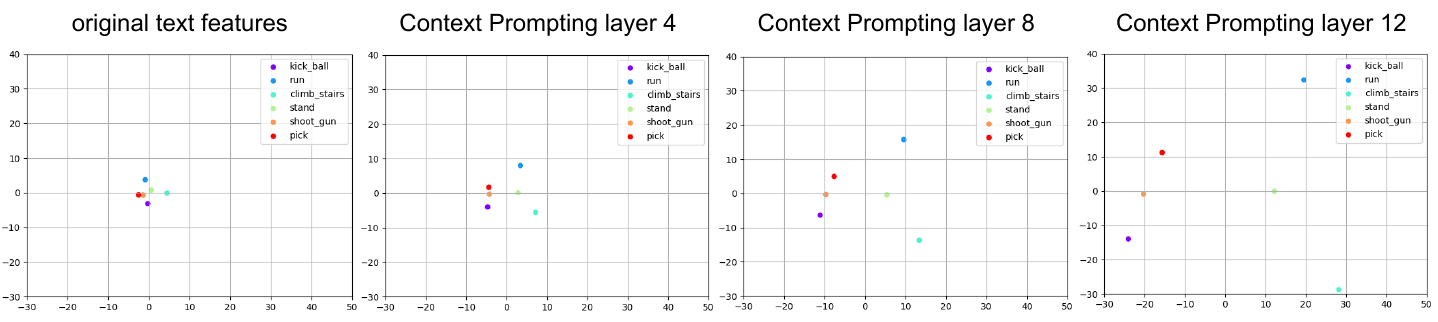}
   \caption{\textbf{Text features distribution prompted by the ”kick ball” visual tokens.}}
\label{fig:jhmdb_label_split_2_kick_ball}
\end{subfigure} \\

\begin{subfigure}{1.0\textwidth}
   \includegraphics[width=1.0\textwidth]{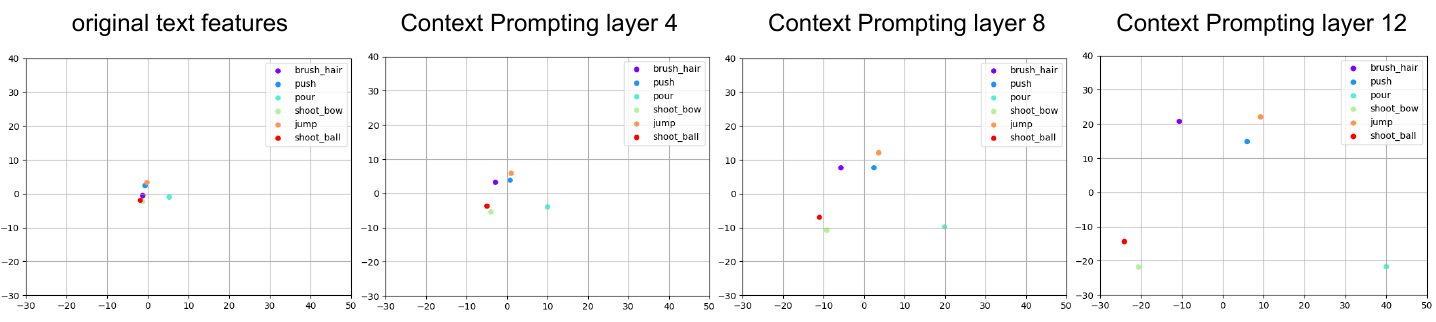}
   \caption{\textbf{Text features distribution prompted by the ”push” visual tokens.}}
\label{fig:jhmdb_label_split_3_push}
\end{subfigure} \\

\begin{subfigure}{1.0\textwidth}
   \includegraphics[width=1.0\textwidth]{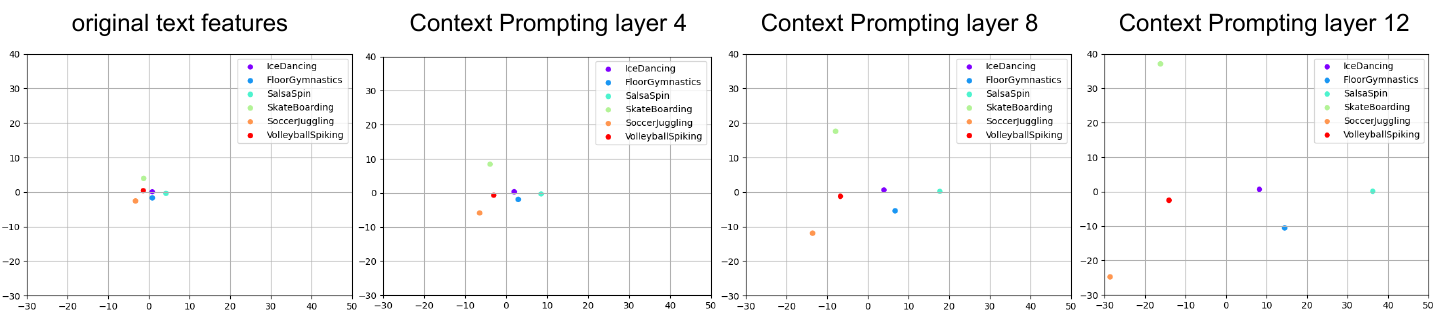}
   \caption{\textbf{Text features distribution prompted by the ”SkateBoarding” visual tokens.}}
\label{fig:ucf_label_split_1_SkateBoarding}
\end{subfigure} \\

\begin{subfigure}{1.0\textwidth}
   \includegraphics[width=1.0\textwidth]{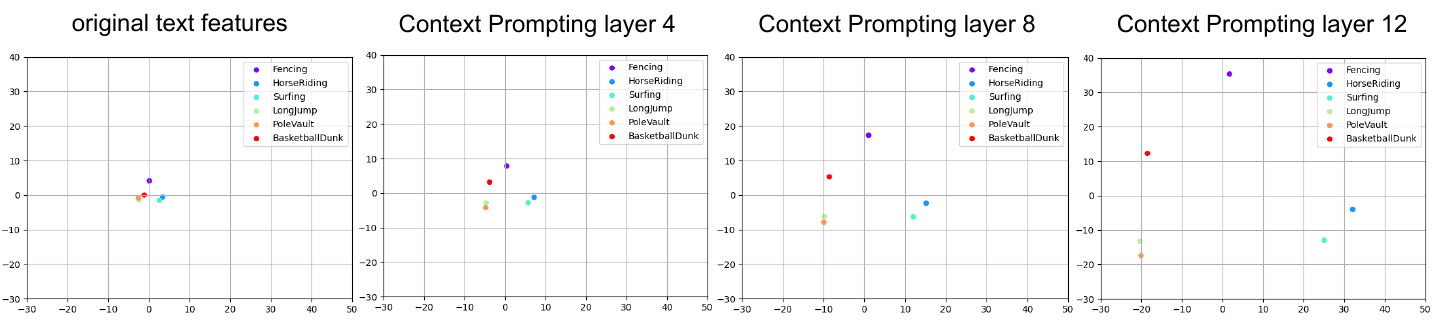}
   \caption{\textbf{Text features distribution prompted by the ”HorseRiding” visual tokens.}}
\label{fig:ucf_label_split_3_HorseRiding}
\end{subfigure}
\caption{\textbf{Text Features Distribution.}}
\label{Fig:Text Features Distribution}
\end{figure*}

As class names inherently carry limited semantic information, relying solely on the text features of these words to calculate similarity with person tokens may introduce ambiguity, potentially impacting the accuracy of action classification. 
To examine the distribution of text features in the feature space, we employ Principal Component Analysis (PCA) to reduce each feature to two dimensions. Subsequently, we illustrate the text features distribution of the CLIP text encoder and various Context Prompting layer outputs in Figure \ref{Fig:Text Features Distribution}.

Initially, the results reveal that regardless of the label group, the original text features derived solely from class names are relatively close, potentially resulting in misclassification of actions. However, our Context Prompting module will utilize different visual information in each layer to augment the text content, thereby gradually increasing the discriminability between each class name. For example, in Figure \ref{fig:ucf_label_split_1_SkateBoarding}, if we rely solely on the original text features to categorize the action "Skateboarding," it may result in misclassification due to the ambiguity with "VolleyballSpiking" in the feature space. However, as these text features extract visual information across multiple prompting layers, the distinction between them becomes more pronounced, which aids in easier differentiation between various actions.

\section{Complexity Analysis}
\label{sec:Complexity Analysis}

\begin{table*}
\begin{adjustbox}{center}
\begin{tabularx}{\textwidth}{XXXXXXr}
    \toprule
    Method & GFLOPs & Training \newline time (s) & Inference \newline time (s) & Throughput \newline (FPS) & Trainable \newline parameters (M) & mAP\\
    \midrule
    baseline & 721.11 & - & 3.95 & 671.14 & - & 64.63\\
    \\
    ActionCLIP \cite{wang2021actionclip} & 409.7 & 199.06 & 9.04 & 293.25 & 18.92 & 69.18\\
    \\
    A5 \cite{ju2022prompting} & 264.15 & 239.34 & 10.33 & 256.63 & 6.35 & 50.92\\
    \\
    X-CLIP \cite{ni2022expanding} & 159.91 & 185.36 & 6.61 & 401.06 & 57.93 & 72.91\\
    \\
    Vita-CLIP \cite{wasim2023vita} & 145.26 & 259.52 & 5.88 & 450.85 & 35.61 & 68.60\\
    \\
    iCLIP \cite{huang2023interaction} & 2410.07 & 2493.1 & 58.11 & 45.62 & 11.6 & 66.53\\
    \\
    \textbf{ST-CLIP (Ours)} & 2431.51 & 806.76 & 31.77 & 83.44 & 55.11 & \textbf{79.62}\\
    \bottomrule
\end{tabularx}
\end{adjustbox}
\caption{\textbf{Complexity analysis on the label split 1 of ZS-JHMDB.} We train/test all the methods on 8 Tesla V100 GPUs.}
\label{tab:complexity}
\end{table*}

We report the complexity analysis on the label split 1 of ZS-JHMDB in Table \ref{tab:complexity}. We calculate the GFLOPs required to infer a video and the training/inference time is reported on all the videos in the train/test split. Compared to \cite{huang2023interaction} and our method, which classifies actions for individuals, these video classification methods only need to classify the entire video once to infer the actions of all detected people in it, potentially resulting in lower GFLOPs and inference time. As for training time, unlike \cite{huang2023interaction} and our method, which optimize on an individual basis during the training phase, these methods optimize on an entire video unit, thus requiring fewer training iterations. However, it is worth mentioning that our method can achieve \textbf{79.62} frame mAP with soft voting, which is a substantial improvement over other methods. Besides, these video classification methods only work on single-action videos, while we can handle videos with multiple actions. Compared to \cite{huang2023interaction}, which also uses individuals as the classification unit, our method requires less training time and achieves 74.55 mAP without soft voting, surpassing \cite{huang2023interaction} by 8.02 mAP.

\section{Additional Experimental Analysis}
\label{sec:Additional Experimental Analysis}
\subsection{Interest Token Spotting on Single-Action Video}

We further conduct experiments applying Interest Token Spotting on single-action video. We present the results on ZS-JHMDB label split 1 with detected boxes and ZS-UCF label split 1 with ground-truth boxes in Table \ref{tab: Interest Token Spotting on Single-Action Video}. The results show that when a video contains only single action, all context tokens can be considered relevant to that action. Therefore, relying solely on the information from interest tokens during the prompting process is insufficient.

\begin{table}[ht]
\begin{adjustbox}{center}
\begin{tabular}[t]{lcc}
    \toprule
    Prompting Tokens & ZS-JHMDB & ZS-UCF \\
    \midrule
    Interest Tokens & 74.16 & 86.50 \\
    All Tokens & \textbf{74.55} & \textbf{87.11} \\
    \bottomrule
  \end{tabular}
\end{adjustbox}
\caption{\textbf{Interest Token Spotting on single-action video}}\label{tab: Interest Token Spotting on Single-Action Video}
\end{table}

\subsection{Complexity Analysis on Proposed Components}

We present the complexity analysis on each of our proposed components in Table \ref{tab: Complexity Analysis on Proposed Components}. The computational complexity (GFLOPs) is calculated on 1 AVA frame with 3 detected people, and we report the trainable parameters with ViT-L backbone. It is worth noting that the majority of Person-Context Interaction's computational complexity arises from using CLIP's pretrained image encoder to extract person features. Additionally, when applying Interest Token Spotting, we must prompt each person’s text features individually, rather than using a shared set of text features, which further increases the computational complexity.

\begin{table}[ht]
\small
\begin{adjustbox}{center}
\begin{tabular}[t]{lcc}
    \toprule
    Components & GFLOPs & Trainable params (M)\\
    \midrule
    Person-Context Interaction & 583.04 & 22.97 \\
    Context Prompting & 21.67 & 189 \\
    Interest Token Spotting & 40.88 & 0 \\
    \bottomrule
  \end{tabular}
\end{adjustbox}
\caption{\textbf{Complexity Analysis on Proposed Components}}\label{tab: Complexity Analysis on Proposed Components}
\end{table}

\subsection{Latency on Multi-Action Video}

We report the latency of our \nickname \space for processing multi-action videos in Table \ref{tab: Latency on Multi-Action Video}. We inference on 1 AVA video which contains 886 frames. Our method takes about 3 minutes for inference when using a single GPU.

\begin{table}[ht]
\begin{adjustbox}{center}
\begin{tabular}[t]{lcc}
    \toprule
    Device & Time (s) & FPS\\
    \midrule
    1 GPU & 190.94 & 4.64 \\
    8 GPUs & 40.22 & 22.03 \\
    \bottomrule
  \end{tabular}
\end{adjustbox}
\caption{\textbf{Latency on 1 AVA video}}\label{tab: Latency on Multi-Action Video}
\end{table}

\subsection{Additional Ablation Study}
We present more ablation studies on the label split 1 of ZS-JHMDB and ZS-AVA.

\noindent\textbf{LoRA in FFN:} In Table \ref{tab:LoRA}, we investigate the impact of LoRA ranks. The results show that when we additionally train learnable matrices with rank 8, we can perform better than relying solely on CLIP's pretrained weight.

\noindent\textbf{Interest tokens:} In table \ref{tab:Interest tokens}, we conduct experiments using different numbers of interest tokens to observe their impact on the results. Our findings indicate that in multi-action videos, introducing too many tokens can potentially sample background noise unrelated to the action, thereby impacting the effectiveness of prompting.

\begin{table}[ht]
\begin{tabularx}{\linewidth}{@{} *{2}{C} @{}}
    \begin{tabular}{@{} lc @{}}
    \toprule
    Rank & mAP \\
    \midrule
    w/o LoRA & 72.80 \\
    r = 2 & 64.52 \\
    r = 4 & 71.70 \\
    r = 8 & \textbf{74.55} \\
    r = 16 & 68.39 \\
    \bottomrule
    \end{tabular}
    \caption{\textbf{LoRA in FFN}}\label{tab:LoRA} & 
    \begin{tabular}{@{} lc @{}}
    \toprule
    Number & mAP \\
    \midrule
    80 & 11.77 \\
    100 & \textbf{12.85} \\
    150 & 12.46 \\
    200 & 11.46 \\
    \bottomrule
    \end{tabular}
    \caption{\textbf{Interest tokens}}\label{tab:Interest tokens}
\end{tabularx}
    \end{table}

\section{Multi-Label Prediction on AVA}\label{sec:Multi-Label Prediction on AVA} In the AVA dataset, each person must perform a pose action, and they may also engage in two additional types of actions: object interaction and person interaction. To handle multi-label prediction, after calculating the cosine similarity using person-context relational tokens and text features, we apply softmax as the activation function for pose actions, and sigmoid for the other two types — object interaction and person interaction.

\section{Results with Groundtruth Bounding Boxes}
\label{sec:Results with Groundtruth Bounding Boxes}
In this section, we present the results using ground-truth bounding boxes of the test data on ZS-AVA, ZS-JHMDB, and ZS-UCF in \cref{tab:ava_evaluation_gt,tab:jhmdb7525_gt,tab:ucf7525_gt} respectively, to analyze the performance of all methods without localization errors.

The results in Table \ref{tab:ava_evaluation_gt} show that our performance is better than most other methods, except for splits 2 and 3, where it is slightly inferior to \cite{wasim2023vita}. However, the performance of video classification methods (e.g., \cite{wasim2023vita}) on multi-action videos largely depends on the quality of tracklets. In the case of using detected boxes to comply with the zero-shot scenario, these methods require additional trackers instead of using ground-truth tracklets, which significantly impacts their performance (avg drops from 10.45 to 3.06). In contrast, our ST-CLIP is not constrained by the tracker, so the performance will not be significantly reduced if we switch to using detected boxes.

\begin{table}[ht]
\begin{adjustbox}{center}
\setlength{\tabcolsep}{1.5mm}
\begin{tabular}{lcccc}
    \toprule
    & \multicolumn{4}{c}{Frame mAP@0.5} \\
    \cmidrule(l){2-5} 
    \multirow{-2}{*}{Method} & split 1 & split 2 & split 3 & avg\\
    \midrule
    baseline & 13.49 & 13.23 & 3.55 & 10.09\\
    baseline (person crop) & 11.88 & 7.06 & 3.72 & 7.55\\
    iCLIP \cite{huang2023interaction} & 5.58 & 12.59 & 2.56 & 6.91\\
    Vita-CLIP \cite{wasim2023vita} & 10.88 & \textbf{14.69} & \textbf{5.78} & 10.45\\
    \textbf{\nickname \space (Ours)} & \textbf{15.80} & 14.32 & 5.16 & \textbf{11.76}\\
    \bottomrule
\end{tabular}
\end{adjustbox}
\caption{\textbf{Evaluation on ZS-AVA.} We report the results using the ground-truth bounding boxes of the test data.}
\label{tab:ava_evaluation_gt}
\end{table}

\begin{table}[ht]
\begin{adjustbox}{center}
\setlength{\tabcolsep}{1mm}
\begin{tabular}{lccccc}
    \toprule
    & \multicolumn{5}{c}{Frame mAP@0.5} \\
    \cmidrule(l){2-6} 
    \multirow{-2}{*}{Method} & split 1 & split 2 & split 3 & split 4 & avg \\
    \midrule
    \multicolumn{6}{l}{\textit{Without the assumption of single-action video}}
    \\
    baseline & 70.17 & 79.91 & \textbf{95.51} & 79.26 & 81.21\\
    ViCLIP \cite{wang2023internvid} & 54.30 & 81.32 & 63.44 & 73.55 & 68.15 \\
    iCLIP \cite{huang2023interaction} & 71.06 & 76.57 & 93.44 & 74.98 & 79.01\\
    \textbf{\nickname \space (Ours)} & \textbf{79.02} & \textbf{82.99} & 94.83 & \textbf{85.28} & \textbf{85.53}\\
    \midrule
    \multicolumn{6}{l}{\textit{With the assumption of single-action video}} \\
    ActionCLIP \cite{wang2021actionclip} & 74.92 & 83.33 & 89.46 & 80.58 & 82.07\\
    A5 \cite{ju2022prompting} & 54.81 & 74.36 & 78.20 & 62.19 & 67.39\\
    X-CLIP \cite{ni2022expanding} & 77.85 & 80.21 & 92.71 & 81.68 & 83.11\\
    Vita-CLIP \cite{wasim2023vita} & 73.13 & \textbf{89.83} & 97.05 & 84.07 & 86.02\\
    \textbf{\nickname \space (Ours)} & \textbf{85.02} & 87.87 & \textbf{97.33} & \textbf{90.27} & \textbf{90.12}\\
    \bottomrule
\end{tabular}
\end{adjustbox}
\caption{\textbf{Evaluation on ZS-JHMDB.} We report the results using the ground-truth bounding boxes of the test data.}
\label{tab:jhmdb7525_gt}
\end{table}

\begin{table}[ht]
\begin{adjustbox}{center}
\setlength{\tabcolsep}{1mm}
\begin{tabular}{lccccc}
    \toprule
    & \multicolumn{5}{c}{Frame mAP@0.5} \\
    \cmidrule(l){2-6} 
    \multirow{-2}{*}{Method} & split 1 & split 2 & split 3 & split 4 & avg\\
    \midrule
    \multicolumn{6}{l}{\textit{Without the assumption of single-action video}} \\
    baseline & 85.70 & 95.11 & 88.17 & \textbf{92.99} & 90.49 \\
    ViCLIP \cite{wang2023internvid} & 75.21 & 74.79 & 44.07 & 58.22 & 63.07 \\
    iCLIP \cite{huang2023interaction} & \textbf{91.30} & 89.91 & 81.56 & 79.12 & 85.47 \\
    \textbf{\nickname \space (Ours)} & 87.11 & \textbf{96.73} & \textbf{91.11} & 92.23 & \textbf{91.80} \\
    \midrule
    \multicolumn{6}{l}{\textit{With the assumption of single-action video}} \\
    ActionCLIP \cite{wang2021actionclip} & 91.33 & 94.39 & 85.26 & 87.90 & 89.72\\
    A5 \cite{ju2022prompting} & 84.74 & 79.67 & 90.00 & 87.98 & 85.60 \\
    X-CLIP \cite{ni2022expanding} & 89.85 & 98.28 & 93.69 & \underline{94.64} & 94.12 \\
    Vita-CLIP \cite{wasim2023vita} & \textbf{94.31} & \textbf{98.80} & \textbf{96.56} & \textbf{96.59} & \textbf{96.57} \\
    \textbf{\nickname \space (Ours)} & \underline{92.90} & \underline{98.75} & \underline{93.97} & 94.44 & \underline{95.02} \\
    \bottomrule
\end{tabular}
\end{adjustbox}
\caption{\textbf{Evaluation on ZS-UCF.} We report the results using the ground-truth bounding boxes of the test data.}
\label{tab:ucf7525_gt}
\end{table}

\cref{tab:jhmdb7525_gt,tab:ucf7525_gt} present the results on ZS-JHMDB and ZS-UCF. On ZS-JHMDB, our method has the best average performance whether using detected boxes or ground-truth boxes. On ZS-UCF, when we use ground-truth boxes to avoid false positives affecting the soft voting results, our method can achieve the second-best performance on most splits and on average.

\section{Limitations}
\label{sec:Limitations}

\begin{figure}
    \begin{center}
   \includegraphics[width=0.5\textwidth]
   {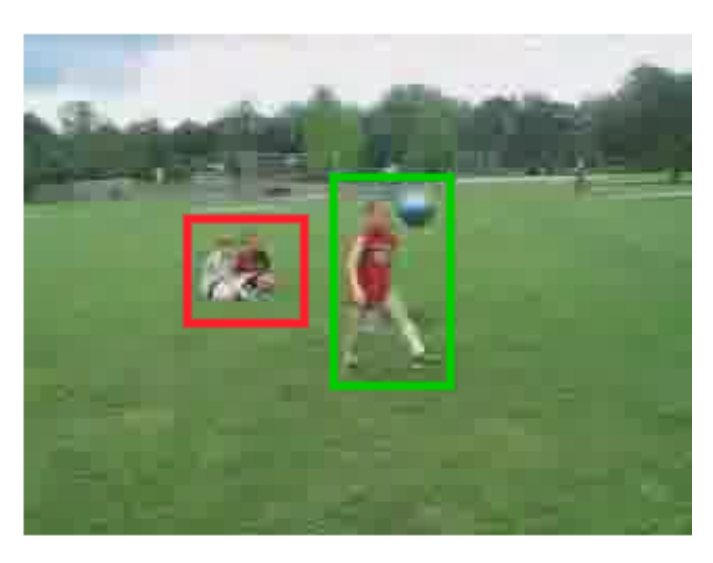}
   \end{center}
   \caption{\textbf{Visualization of wrong detected person bounding box.} The image is from the test data of class "SoccerJuggling", and the boxes are detected by Faster R-CNN. The red box is counted as a false positive according to the labeling in this dataset since the people inside are not doing the "SoccerJuggling" action.}
\label{fig:wrong_detected_box}
\end{figure}

In certain scenarios, there may be unrelated individuals who are not engaged in any actions. In such cases, these individuals should be treated as background elements. However, achieving this necessitates the human detector to learn from samples of a specific action class to effectively detect only the person executing this action. In a zero-shot setting, the person detector is prone to experiencing more localization errors as it has not been exposed to samples of the testing classes. Taking Figure \ref{fig:wrong_detected_box} as an example, the Faster R-CNN may detect two person bounding boxes with high confidence scores, although only one of them is genuinely performing the action "SoccerJuggling". Consequently, the detection of the other box will be considered a false positive. 

Given our method's two-stage pipeline nature, the aforementioned localization error will influence our performance in two aspects: (1) In processing single-action videos, these false positive instances will contribute to soft voting, thereby compromising the classification performance to some extent. (2) When handling multi-action videos, the person tokens from these incorrectly detected boxes will be employed for interaction modeling, which will influence the effectiveness of the person-person interaction involving the target individual.

\section{Implementation Details}
\label{sec:Implementation Details}
In this section, we provide the implementation details for each compared method in our experiments. For iCLIP \cite{huang2023interaction}, we employ the best-performing model featuring four types of interaction modules and Interaction-Aware Prompting. For ActionCLIP \cite{wang2021actionclip}, to follow their approach, we sample 8 frames from each video, then utilize 6 transformer encoder layers to perform temporal modeling. Besides, we adopt the handcrafted prompts they proposed to prompt labels. For Efficient-Prompting \cite{ju2022prompting}, we sample 16 frames from each video, and employ the A5 model they proposed, which uses 2 encoder layers for temporal modeling and prepends/appends 16 vectors with the textual embeddings. For X-CLIP \cite{ni2022expanding}, we sample 8 frames from each video, and use the proposed Cross-frame Communication and 1-layer Multi-frame Integration Transformer to generate video features. Besides, we also leverage the Video-specific Prompting in their method to prompt labels. For Vita-CLIP \cite{wasim2023vita}, we sample 8 frames from each video, and utilize 8 Global Video-Level Prompts following their method. For ViCLIP \cite{wang2023internvid}, we sample 8 frames from each video and use the video encoder to obtain the video feature map. Then, based on each person's bounding box, we apply ROIAlign to extract individual person features for classification.
As for the training iterations, for iCLIP, we train the network for 7K iterations on ZS-JHMDB and 10K iterations on ZS-UCF as in their method. We train other video classification methods for 1K iterations on ZS-JHMDB and ZS-UCF, and more iterations will lead to lower performance due to overfitting. For all the aforementioned methods except ViCLIP \cite{wang2023internvid}, we utilize the CLIP backbone and freeze the image and text encoder, consistent with our approach. As for ViCLIP \cite{wang2023internvid}, we freeze the text encoder and finetune the video encoder during training. The video encoder is pretrained on the InternVid-10M-FLT \cite{wang2023internvid} dataset. We provide the details of training hyperparameters for all methods on ZS-AVA, ZS-JHMDB, and ZS-UCF in \cref{tab:ava_hyperparameters,tab:jhmdb_hyperparameters,tab:ucf_hyperparameters}.

\begin{table*}
\vspace{-2cm}
\begin{adjustbox}{center}
\setlength{\tabcolsep}{2mm}
\begin{tabular}{lcccccc}
    \toprule
     Method & Iterations & Learning rate & Warmup iterations & Warmup factor & Optimizer & Batch size\\
    \midrule
    iCLIP \cite{huang2023interaction} & 30000 &  & 2000 &  &  & \\
    Vita-CLIP \cite{wasim2023vita} & 10000 & 0.0004 & 1000 & 0.25 & SGD & 8\\
    \textbf{\nickname \space (Ours)} & 20000 &  & 2000 &  &  & \\
    \bottomrule
\end{tabular}
\end{adjustbox}
\caption{\textbf{hyperparameters on ZS-AVA.} All methods employ ViT-L/14 backbone.}
\label{tab:ava_hyperparameters}
\end{table*}

\begin{table*}
\vspace{-8cm}
\begin{adjustbox}{center}
\setlength{\tabcolsep}{2mm}
\begin{tabular}{lcccccc}
    \toprule
     Method & Iterations & Learning rate & Warmup iterations & Warmup factor & Optimizer & Batch size\\
    \midrule
    ActionCLIP \cite{wang2021actionclip} &  &  &  &  &  &  \\ 
    A5 \cite{ju2022prompting} &  &  &  &  &  & \\
    X-CLIP \cite{ni2022expanding} & \multirow{-1}{*}{1000} & \multirow{-1}{*}{0.0002} & \multirow{-1}{*}{700} & \multirow{-1}{*}{0.25} & \multirow{-1}{*}{SGD} & \multirow{-1}{*}{8}\\
    Vita-CLIP \cite{wasim2023vita} &  &  &  &  &  & \\
    ViCLIP \cite{wang2023internvid} &  &  &  &  &  & \\
    \midrule
    iCLIP \cite{huang2023interaction} & 7000 & 0.0002 & 700 &  &  & \\
    \textbf{\nickname \space (Ours)} & 3000 & 0.00025 & 800 & \multirow{-2}{*}{0.25} & \multirow{-2}{*}{SGD} & \multirow{-2}{*}{8} \\
    \bottomrule
\end{tabular}
\end{adjustbox}
\caption{\textbf{hyperparameters on ZS-JHMDB.} All methods employ ViT-B/16 backbone.}
\label{tab:jhmdb_hyperparameters}
\end{table*}

\begin{table*}
\vspace{-8cm}
\begin{adjustbox}{center}
\setlength{\tabcolsep}{2mm}
\begin{tabular}{lcccccc}
    \toprule
     Method & Iterations & Learning rate & Warmup iterations & Warmup factor & Optimizer & Batch size\\
    \midrule
    ActionCLIP \cite{wang2021actionclip} &  &  &  &  &  &  \\ 
    A5 \cite{ju2022prompting} &  &  &  &  &  & \\
    X-CLIP \cite{ni2022expanding} & \multirow{-1}{*}{1000} & \multirow{-1}{*}{0.0002} & \multirow{-1}{*}{1000} & \multirow{-1}{*}{0.25} & \multirow{-1}{*}{SGD} & \multirow{-1}{*}{8}\\
    Vita-CLIP \cite{wasim2023vita} &  &  &  &  &  & \\
    ViCLIP \cite{wang2023internvid} &  &  &  &  &  & \\
    \midrule
    iCLIP \cite{huang2023interaction} & 10000 & 0.0002 & 1000 &  &  & \\
    \textbf{\nickname \space (Ours)} & 3000 & 0.00025 & 800 & \multirow{-2}{*}{0.25} & \multirow{-2}{*}{SGD} & \multirow{-2}{*}{8} \\
    \bottomrule
\end{tabular}
\end{adjustbox}
\caption{\textbf{hyperparameters on ZS-UCF.} All methods employ ViT-B/16 backbone.}
\label{tab:ucf_hyperparameters}
\end{table*}

\clearpage
{\small
\bibliographystyle{ieee_fullname}
\bibliography{main}
}

\end{document}